\documentclass{article}[12pt]
\usepackage{geometry,amsthm,amsmath,amssymb,graphicx,float,enumerate,hyperref}
\usepackage{subfigure}
\usepackage{arydshln}
\usepackage{pgf,tikz}
\usepackage{mathrsfs}
\usepackage[font=small]{caption}
\usepackage[margin=0.35in]{caption}

\usepackage{mathpazo}
 \usepackage{amsmath, amsthm}

\newcommand{\R}{\mathbb{R}}

\newtheorem{theorem*}{Theorem}

\newtheorem{definition*}{Definition}

\newtheorem{lemma*}{Lemma}

\newtheorem{assumption*}{Assumption}

\newtheorem{corollary*}{Corollary}
\newtheorem{problem*}{Problem}

\newtheorem{observation*}{Observation}
\newtheorem{hypothesis*}{Hypothesis}

\title{Understanding robustness and generalization of \\ artificial neural networks through Fourier masks.}

\author{Nikos Karantzas$^{1,3\dagger*}$, Emma Besier$^{1,3\dagger}$, Josue Ortega Caro$^{1,3}$, Xaq Pitkow$^{1,2,3}$, \\ Andreas S. Tolias$^{1,2,3}$, Ankit B. Patel$^{1,2,3}$ and Fabio Anselmi$^{1,3,*}$}

\date{
{\small $^{1}$Department of Neuroscience, Baylor College of Medicine, Houston, 77030, USA.}\\
{\small $^{2}$Department of Electrical and Computer Engineering, Rice University, Houston, 77005, USA.} \\
{\small $^{3}$Center for Neuroscience and Artificial Intelligence, Baylor College of Medicine, Houston, 77030, USA.}\\
{\small$\dagger$  co-first authors} \\
{\small$*$ corresponding authors \footnote{fabio.anselmi@bcm.edu, anselmi@mit.edu, nikos.karantzas@bcm.edu}}
}

\begin{document}
\maketitle
\abstract 
\noindent Despite the enormous success of artificial neural networks (ANNs) in many disciplines, the characterization of their computations and the origin of key properties such as generalization and robustness remain open questions. Recent literature suggests that robust networks with good generalization properties tend to be biased towards processing low frequencies in images. To explore the frequency bias hypothesis further, we develop an algorithm that allows us to learn \emph{modulatory masks} highlighting the \emph{essential input frequencies} needed for preserving a trained network's performance. We achieve this by imposing \emph{invariance} in the loss with respect to such modulations in the input frequencies. We first use our method to test the low-frequency preference hypothesis of adversarially trained or data-augmented networks. Our results suggest that adversarially robust networks indeed exhibit a low-frequency bias but we find this bias is also dependent on directions in frequency space. However, this is not necessarily true for other types of data augmentation. Our results also indicate that the essential frequencies in question are effectively the ones used to achieve generalization in the first place. Surprisingly, images seen through these modulatory masks are not recognizable and resemble texture-like patterns.


\section{Introduction}
Artificial neural networks (ANNs) have achieved impressive performance in a variety of tasks, e.g., object recognition, function approximation, natural language processing, etc. [\cite{HintonNature}]. However, their computational capacity remains rather opaque. In particular, the operations performed by ANNs are profoundly constrained by the choice of architecture, initialization, optimization techniques, etc., and such constraints have a significant impact on key properties such as generalization power and robustness. Studying adversarial robustness has been a very active area of research, since it is closely related to how trustworthy and reliable neural networks can be [\cite{goodfellow2014explaining}]. One of the most explored directions has been the analysis of adversarial perturbations from a frequency standpoint. For example, the work of [\cite{yin2019fourier}] establishes a relationship between the frequency domain of different noises (e.g Adversarial examples and Common corruptions) and model performance. In particular, they show that deep neural networks are more sensitive to high frequency adversarial attacks or common corruptions such as random noise, contrast change, and blurring. Additionally, adversarial perturbations of commonly trained models tend to be higher frequency than their adversarially trained counterparts. Furthermore, \cite{wang2020frequencybased} found that high frequency features are necessary for good generalization performance while the work of \cite{SharmaIJCAI2019} shows that performance improvements in white-box and black-box transfer settings can be achieved only when low frequency components are preserved.

These results have led to various methodologies that help us understand artificial neural networks through a frequency lens. One such method is Neural Anisotropic Directions (NADs) [\cite{ortiz2020hold,ortiz2020neural}]. NADs are input directions for which a network is able to linearly classify data. Furthermore, \cite{tsuzuku2019structural} introduced a method to compute a neural network's sensitivity to input directions in the Fourier domain. Moreover, \cite{Li2022} show that robust deep learning object recognition models rely on low frequency information in natural images. Finally \cite{Abello} divides the image frequency spectrum into disjoint disks and provides  evidence that mid or high-level frequencies are important for ANN classification.

In this work we introduce a simple and easy-to-use method to \textit{learn} the input frequency features that a network deems essential in order to achieve its classification performance. We visualize the relevant frequencies by learning a \textit{modulatory mask} on the Fourier transform of the input data that defines a modulation-invariant loss function obtained via a simple optimization algorithm (Section \ref{sec:approach}). We compare such masks with their adversarially trained  or data augmented counterparts (Section \ref{sec:dataaug}). In the case of adversarial training, the comparison is done at two levels of analysis. At a global level, we learn a mask for the entire test set. Our goal is to find the frequencies that allow for \textit{robust generalization}. At a single image level, we explore the frequencies responsible for adversarial success/failure. Those comparisons allow us to test the hypothesis that adversarially trained models have a bias towards low frequency features and assess if the same holds for other types of data augmentation. 

In the case of adversarial augmentation, our results confirm the low frequency bias hypothesis. However, they also highlight that the important frequency redistribution due to the augmentation is highly anisotropic. In the case of common data augmentations instead, our results show how the frequency reorganization depends on the type of augmentation, e.g., rotation- or scale-augmented models exhibit mid-high and low frequency biases, respectively.   

The single-image mask analysis reveals that only a few, class-specific frequencies are crucial to determine a network's decision. Moreover, \textit{those frequencies are effectively the ones used to achieve its performance}. In fact, mask-filtered images do not alter performance at all. However, surprisingly, they are not recognizable. They are characterized by texture-like patterns. This is in line with previous work by  [\cite{Geirhos2019}], which provided evidence that Convolutional Neural Networks (CNNs) are biased towards textures rather than shapes in object recognition. Our method differs from all previous ones in that we explicitly learn the frequencies defining the features a model is sensitive to.

\section{Methods}\label{sec:methods}
\subsection{Approach}\label{sec:approach}
\begin{figure}[h!]
\centering
\includegraphics[scale=0.25]{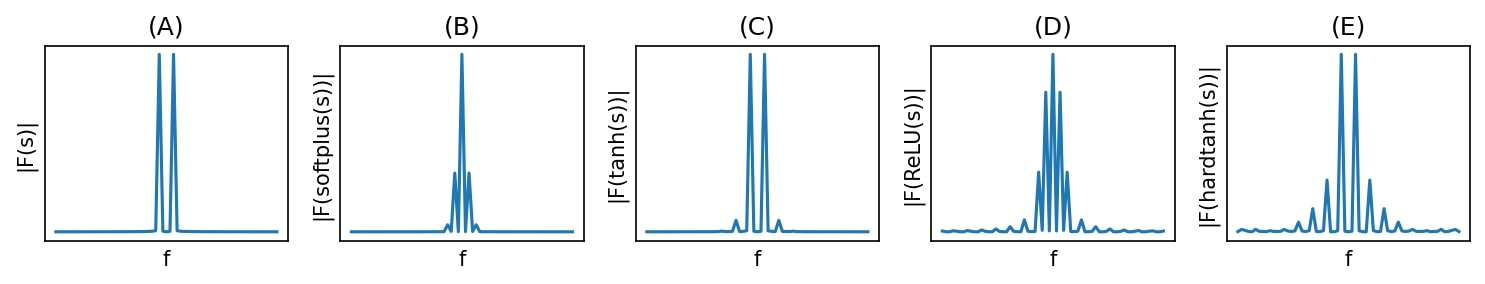}
\caption{Non-linear distortions in the frequency domain due to the application of (B) softplus, (C) tanh, (D) ReLU and (E) hardtanh non-linear activations on $s(\cdot)=\sin(\cdot)$ (A).}
\label{fig:nlsimple}
\end{figure}

Artificial neural networks and their associated task-dependent losses define highly non-linear functions of their input. In terms of the frequency content found in a signal, the effect of the application of a non-linear function can be understood by considering the following simple one-dimensional example. Suppose $f(t)= \cos(w_{1}t)+\cos(w_{2}t)$ is a sound wave and let $\sigma(t)=t^{2}$. Then 
\[
(\sigma\circ f)(t) = \frac{1}{2}[2+\cos(2w_{1}t)+\cos(2w_{2}t)+2\cos((w_{1}+w_{2})t)+2\cos((w_{1}-w_{2})t)].
\] 
We see that one of the effects of $\sigma$ on $f$ is to generate the new frequency components $w_1 - w_2$, $w_1 + w_2$, $2w_{1}$, $2w_{2}$. The first two are due to a phenomenon called \textit{intermodulation}, the last are due to what is called \textit{harmonic distortion}. Harmonic distortion has been studied in the context of neural networks with different activation functions by [\cite{mehmeti2021ringing}], where an empirical demonstration and theoretical arguments are given to support the claim that the presence of non-linear elements mainly causes a spread in the frequency content of the loss function. Their reasoning is the following: let $\phi:\R\to\R$ be a non-linear function and $T\phi$ denote its Taylor expansion around the origin. For $x\in \R^{d}$, using the convolution theorem yields
\begin{equation}\label{eq:selfconv}
FT\phi(x)= F\sum_n a_{n}\underbrace{x\odot\cdots\odot x}_\textrm{n-times} = \sum_n a_{n}\underbrace{\hat{x}*\cdots*\hat{x}}_\textrm{n-times},
\end{equation}
where $\phi$ is acting pointwise on the components of $x$, $Fx=\hat{x}$, and the RHS is a weighted sum of self-convolutions. [\cite{mehmeti2021ringing}] show that repeated convolutions broaden the frequency spectrum by adding higher frequency components corresponding to large coefficients $a_{n}$, an effect they call ``blue shift". A visual illustration of the blue-shift effect is shown in Figure \ref{fig:nlsimple} where we considered a one dimensional sinusoidal stimulus $s$ filtered by softplus, tanh, ReLU, and hardtanh non-linearities. Additional to the blue shift effect (harmonic distortion), we also see the impact of intermodulation. 

Let us now consider a more complex non-linear function such as a trained neural network. In this case, the non-linear distortion induced by the network will be manifested in its representation space and therefore in its decision making.

\begin{figure}[h!]\centering
\includegraphics[width=0.7\textwidth]{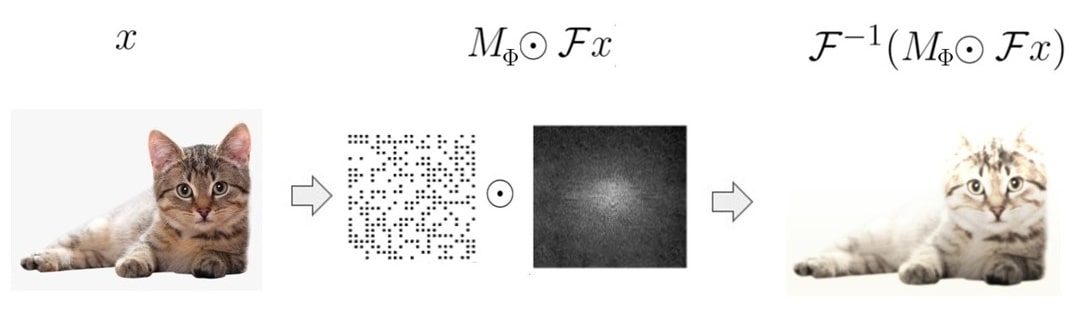}
\caption{A schematic of the preprocessing layer defined by the mask: the input $x$ is transformed into the Fourier domain where it is filtered with a learnable mask $M_{\Phi}$. $M_{\Phi}\odot \mathcal{F}x$ is then remapped into the pixel domain through the inverse Fourier transform $\bar{x} = \mathcal{F}^{-1}(M_{\Phi}\odot\mathcal{F}x)$.}
\label{fig:1}
\end{figure}

As mentioned above, one of the purposes of this work is to propose an algorithm to identify the \textit{essential input frequencies in a trained ANN's decisions}. To this end, let us consider an image dataset $\mathcal{X} = \{(x_{i},y_{i})\}_{i=1}^{N}$, where $x_{i}\in\R^{d\times d}$ denotes the $i$-th input image and $y_{i}\in\mathbb{Z}_{C}$ its associated label ($C$ denotes the number of classes). We split $\mathcal{X}$ into a training set $\mathcal{X}_{T}$ and a validation set $\mathcal{X}_{V}$. We obtain the masks via the following optimization algorithm: we first pre-train a network $\Phi$ on $\mathcal{X}_{T}$ with the objective of solving a classification task. We subsequently freeze the weights of $\Phi$ and attach a pre-processing layer whose weights are the entries $m_{ij}$ of a mask matrix $M_{\Phi}\in\mathbb{R}^{d\times d}$. This layer acts as follows: for every $x\in \mathcal{X}_{V}$ we modulate its Fourier transform $\mathcal{F}x$ by computing the product $M_{\Phi}\odot \mathcal{F}x$, where $\odot$ indicates the Hadamard product. We next compute the inverse Fourier transform $\bar{x} = \mathcal{F}^{-1}(M_{\Phi}\odot\mathcal{F}x)$, which is then fed into the network (see Figure \ref{fig:1}). 
Finally, we learn the mask $M_{\Phi}$ by solving the optimization problem
\begin{equation}
\label{min_problem}
M_{\Phi}(\lambda, p)=\text{argmin}_{M_{\Phi}}\sum_{x \in \mathcal{X}_{V} }e^{\left[\mathcal{L}(\Phi(\bar{x}), y)-\mathcal{L}(\Phi(x), y)\right]^{2}}+\lambda \|M_{\Phi}\|_{p},\quad\lambda\in \R_{+},
\end{equation}
where $\Phi$ denotes the  pre-trained network, $\lambda\|M_{\Phi}\|_{p}$ is a regularization term penalizing the $p$-norm of the learned mask, and $\mathcal{L}$ is the loss function associated with the classification task. The first term in Equation \eqref{min_problem} enforces an \textit{invariance} in the loss with respect to the transformation $x\mapsto\bar{x}$ induced by the mask. The latter
is key because we are expecting the desired frequencies to be revealed when there is no change in the loss $\mathcal{L}$ and maximal change in the $p$-norm of the mask $M_{\Phi}$. In other words, the mask is determined by a \textit{symmetry operation in the Fourier space of the input with minimal $p$-norm}. A solution to Equation \eqref{min_problem} is a mask $M_{\Phi}$ addressing the question: which frequencies are essential in this trained ANN's decision making? Such masks, obtained for various data augmentation choices reveal the frequencies associated with each particular choice.

At this point, we note that the mask is learned on the validation set $\mathcal{X}_{V}$ and not on the training set $\mathcal{X}_{T}$. This is because we are interested in exploring the minimal set of frequencies preserving \textit{generalization} power of $\Phi$. Moreover, we tested the stability of our mask generation algorithm across different runs. This is crucial since it attests to the reliability of the qualitative and quantitative analysis. We also note that masks can be obtained for a single images, simply considering a single $x \in \mathcal{X}_{V}$ in Equation \eqref{min_problem} instead of the full validation set.

\subsection{Dataset and simulations}\label{sec:ds}
Our data consisted of $6,644$ image/label pairs from $5$ classes of ImageNet [\cite{Imagenet}]. $4,710$ of those pairs belong to our training set $\mathcal{X}_{T}$ and the remaining $1,934$ pairs belong to our validation set $\mathcal{X}_{V}$. For simplicity, we choose grayscale versions of our dataset images, though our method can be applied for any number of input channels. Our images were centered with respect to the mean and standard deviation of $\mathcal{X}_{T}$.

We initially trained a VGG11 [\cite{vgg}] baseline model on $\mathcal{X}_{T}$ using the Pytorch framework. For each subsequent training run we varied the type of data augmentation used for pre-processing (adversarial examples, random scales, random translations, random rotations).

Each of the $5$ networks in total was trained using the Adam optimizer ([\cite{Adam}]) and a maximum learning rate of $10^{-3}$. The learning rate of each learnable parameter group was scheduled according to the one-cycle learning rate policy with a minimum value of $0$ ([\cite{Smith2017CyclicalLR}]). We found that this set of hyperparameter choices allowed us to achieve stable training for all our models. We trained each model for a maximum of $50$ epochs and eventually evaluated our models on the validation set $\mathcal{X}_{V}$. We finally saved the weight-state of each model that achieved the minimum Cross Entropy loss within the chosen interval of epochs. For each of our pre-trained networks, we learn its corresponding Fourier mask according to the algorithmic process presented in section \ref{sec:approach}. We use $\ell_{1}$-regularization on the norm of the mask to enforce sparsity. In the next section we present masks for every data augmentation scheme we chose as well as their respective differences. For a given set of masks, we center the mask differences around the origin. This helps with the interpretation of the masks without altering the geometry of the particular set.   

\section{Results}\label{sec:dataaug}
Adversarial training can be seen as a type of data augmentation where the inputs are augmented with adversarial examples [\cite{goodfellow2014explaining}] to increase robustness to adversarial attacks. Here we test the commonly accepted hypothesis that adversarially trained models need low frequency features for robustness. We do so by comparing the Fourier mask learned for a vanilla network $\Phi_{N}$ with that of an adversarially trained network $\Phi_{A}$ when the learning occurs over the whole validation set. Specifically, we compare a naturally trained VGG11 with an adversarially trained one using the \emph{torchattacks} library [\cite{torchattacks}] and a Projected Gradient Descent attack (PGD). [\cite{caro2020local}] has shown the frequency structures of adversarial attacks are similar across different adversarial attacks. Therefore, although the set of potential choices one can explore is vast, in this work we focus on PGD for simplicity. 
Besides the mask difference we also compute the radial and angular energy of each mask by considering radial and angular partitions of the frequency domain (Figure \ref{fig:all_energy} (A), (B)).
We then test if the same low-frequency preference hypothesis holds true in the case of common data augmentations. To gain some intuition, let us consider a simple one layer network whose representation is given by $\Phi(x)=\sigma \langle w,x\rangle$, where $\sigma:\R\to\R$ is a non-linear function, $x,w\in \R^{d}$, and $\ell:\R\to\R_{+}$ is a cost function. We consider data augmentations generated by a group of transformations $G := \{g_{\theta}: \theta\in\R\} \subset \R^{d\times d}$. The augmented loss can now be expressed as
\[
\mathcal{L}(w) = \frac{1}{N} \sum_{i=1}^{N} \int \ell\left(\sigma \langle w, g_{\theta}x_{i} \rangle; y_{i}\right) d\theta = \frac{1}{N} \sum_{i=1}^{N} \int \ell\left(\sigma \langle g^{*}_{\theta}w, x_{i} \rangle; y_{i}\right) d\theta, \quad (x_{i}, y_{i}) \in \mathcal{X},
\]
where the second equality holds because $\langle w, g_{\theta}x_{i}\rangle=\langle g^{*}_{\theta}w, x_{i}\rangle$ and $g^{*}$ denotes the adjoint. We note that in this context the loss function is \textit{invariant to $G$ transformations of the weights}, i.e., $\mathcal{L}(g_{\theta}w)=\mathcal{L}(w)$ for any $g_{\theta}\in G$ (the proof of this statement relies on simple properties of group transformations, see [\cite{Dobriban2020}]). Here we explore the impact such an invariance of the loss function has on the learned Fourier masks. The reasoning is as follows: updating the weights of an ANN is achieved through gradient descent, i.e., $\Delta w_{t}=-\alpha \nabla_{w} \mathcal{L}(w_{t})$, where $w_{t}$ denotes the weights of the network at iteration $t$ and $\alpha\in\R^{+}$ is the learning rate. The frequency content of the gradient of the loss at iteration $t$ affects the frequency content of the weights. In turn, the latter determine the input frequencies the network is analyzing and thus will determine the mask. In other words, the frequency content of the loss, as well as how it is modified by different data augmentations, will impact the frequency content observed in the mask. 

Let us consider a simple one dimensional example ($d=1$) and the translation operator. In this case the loss $\mathcal{L}$ is \textit{invariant to translations of the weights}, i.e., 
\[
\mathcal{L}(T_{t}(w))=\mathcal{L}(w), \quad \forall t\in\R,
\] 
where $T_{t}: \R \to \R$ is the translation operator defined as $T_{t}(\cdot) = \cdot - t$. For $x_{i}\in\mathcal{X}$ and $t\in \R$, let $q_{i}(\cdot) := \ell\left(\sigma(T_{t}(\cdot) x_{i}); y_{i}\right)$. Then the Fourier transform of $\mathcal{L}$ yields
\[
\mathcal{F}(\mathcal{L})(\gamma) = \frac{1}{N} \sum_{i=1}^{N} \int_{-\infty}^{+\infty} \mathcal{F}(q_{i})(\gamma) e^{-2\pi i \gamma t} dt = \frac{1}{N} \sum_{i=1}^{N} \delta(\gamma) \mathcal{F}(q_{i})(\gamma) = \frac{1}{N} \sum_{i=1}^{N} \mathcal{F}(q_{i})(0)
\]
where we used the translation property of the Fourier transform and $\delta$ denotes the Dirac delta. This simple example illustrates the effect of the translation operator on the loss $\mathcal{L}$, i.e., a shift towards low frequencies (in this case a full shift of all frequencies to the DC component, the only non-zero component in the above equation). Note that an augmentation with all possible translations is not realistic. However, even a finite range of translations in the interval $t\in [-a, a]$, for a sufficiently large $a$, will produce a similar effect. Indeed, we have
\[
\mathcal{F}(\mathcal{L})(\gamma) = \frac{1}{N} \sum_{i=1}^{N} \int_{-\infty}^{\infty} \mathcal{F}(q_{i})(k) \chi_{[-a,a]}(t) e^{-2\pi i\gamma t}dt = \frac{2a}{N} \sum_{i=1}^{N}
\text{sinc}(2\pi\gamma a) \mathcal{F}(q_{i})(\gamma)
\]
where $\chi$ denotes the characteristic function. Thus, the impact of averaging over an interval of translations on $\mathcal{L}$ is to dampen its frequencies with a sinc function profile, i.e., a frequency re-weighting with a \textit{bias for low frequencies}. However, we stress that the above argument is developed with a $1$-layer network in mind. The effect of data-augmentation with respect to random translations viewed through a deep network is expected to be more intricate.

\subsection{Masks generated for the whole dataset}

\begin{figure}[h!]\centering
\includegraphics[width=0.9\textwidth]{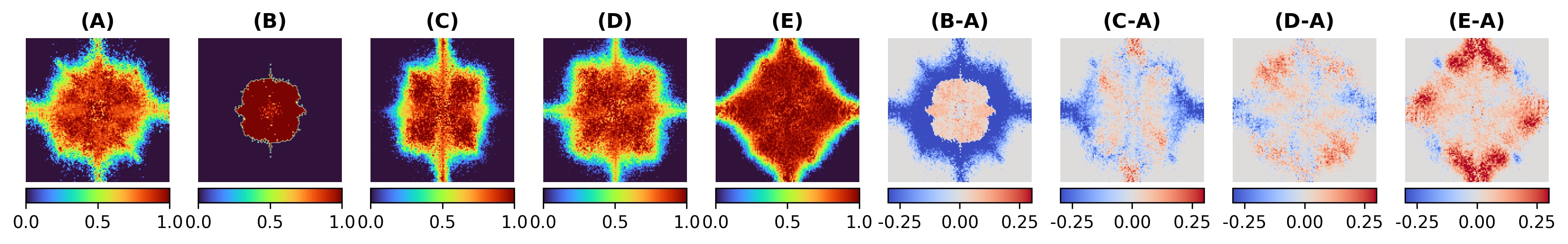}
\includegraphics[width=0.5\linewidth]{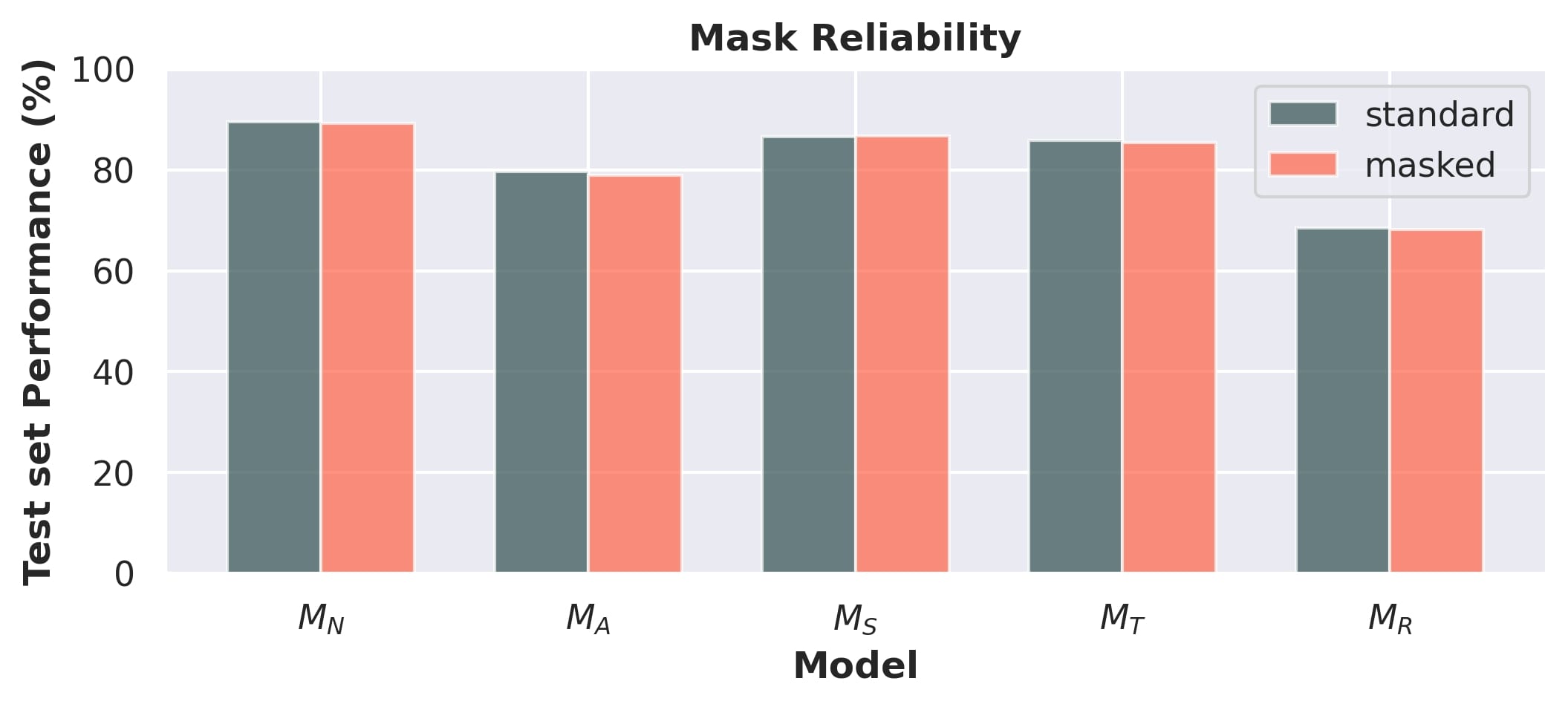}
\caption{\textit{Top}: Learned masks for (A) vanilla network ($M_{N}$), (B) Adversarially trained ($M_{A}$) (C) scale-invariant network ($M_{S}$), (D) translation-invariant network ($M_{T}$), (E) rotation-invariant network ($M_{R}$) and their differences. \textit{Bottom}: Model performance with and without the mask layer.}
\label{fig:dataaug_vgg}
\end{figure}

\begin{figure}[h!]\centering
\includegraphics[width=0.85\textwidth]{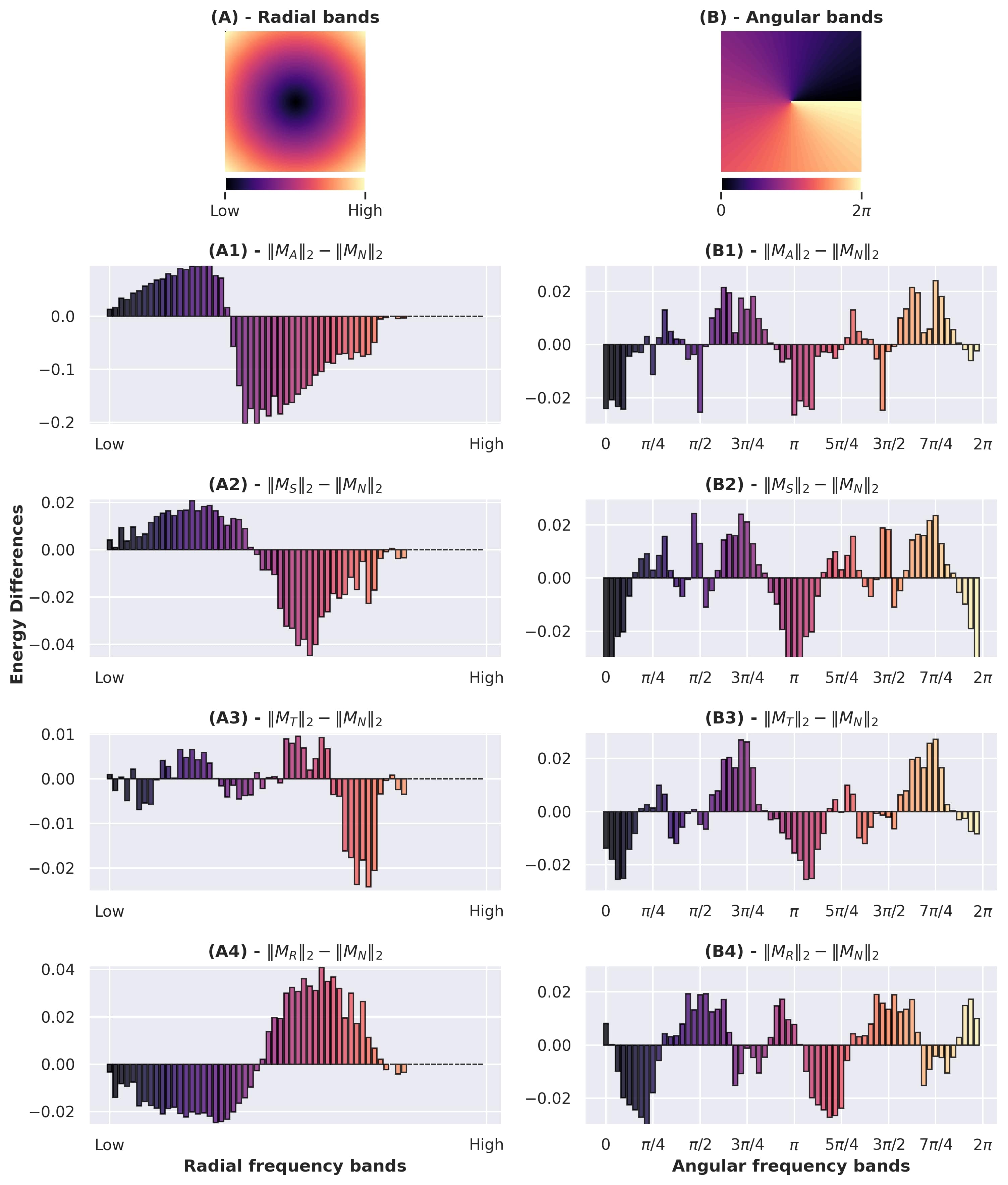}
\caption{(A) Radial Partitions of the Fourier domain (B) Angular partitions of the Fourier domain; Energy difference in radial (A1, A2, A3, A4) and angular (B1, B2, B3, B4) directions, for the augmentations in $B, C, D, E$ of Figure \ref{fig:dataaug_vgg} ($\|M_{i}\|_2-\|M_{N}\|_2, i=$ adversarial (A), scale (S), translation (T) and rotation (R) augmentations).}
\label{fig:all_energy}
\end{figure}

We generated masks over $\mathcal{X}_{V}$ for  networks trained to be robust to adversarial examples, random scales, translations, and rotations. The masks in Figure \ref{fig:dataaug_vgg}(Top) and their differences reveal how distinct frequency biases depend on the type of data augmentation. We also note how model performance is not altered by the introduction of the mask layer Figure \ref{fig:dataaug_vgg}(Bottom).
In the case of adversarial augmentation there exists a net bias towards low frequencies as shown by the difference between the masks generated by the vanilla and adversarial trained network in Figure \ref{fig:dataaug_vgg} (B-A). This is further confirmed by the radial energy difference in Figure \ref{fig:all_energy}(A1), while the angular energy difference in Figure \ref{fig:all_energy}(B1) shows that the redistribution of the frequencies occurs anisotropically.

In the case of common augmentations our results exhibit contrasting effects in the Fourier masks. While the redistribution of the mask frequencies seems to be directionally-dependent (Figure \ref{fig:all_energy} B1, B2, B3), only robustness to scales endows the net with a bias towards low frequencies (Figure \ref{fig:all_energy} (A2)). For translations the mask implies a less clear effect (Figure \ref{fig:all_energy} (A3)), where a mixed behavior is present for mid and low frequencies. Interestingly, in the case of rotational robustness, Figure \ref{fig:all_energy}(A4) shows a high frequency bias.   

\subsection{Masks generated for single images}

\begin{figure}[ht]\centering
\includegraphics[width=0.9\textwidth]{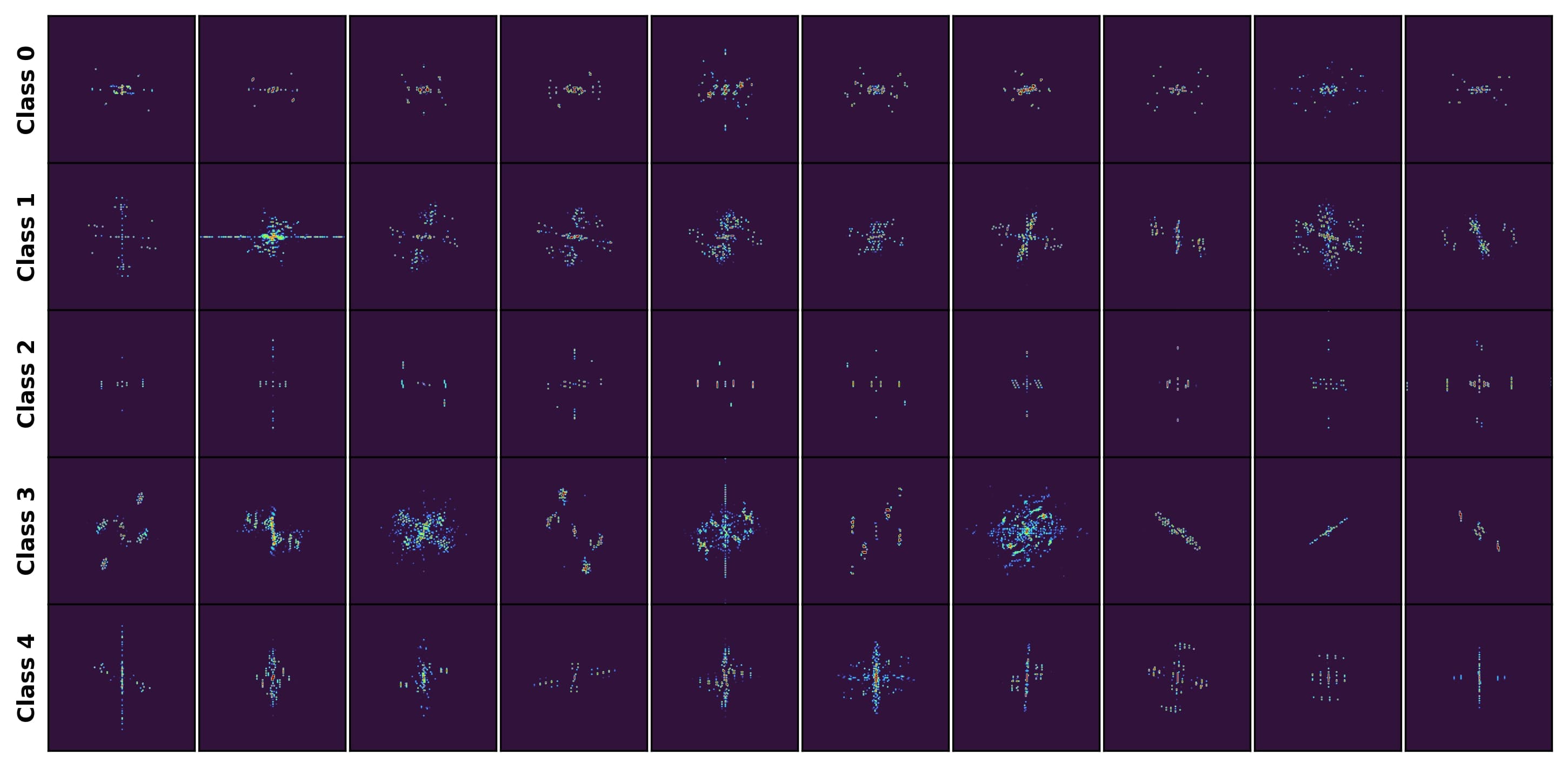}
\includegraphics[width=0.9\textwidth]{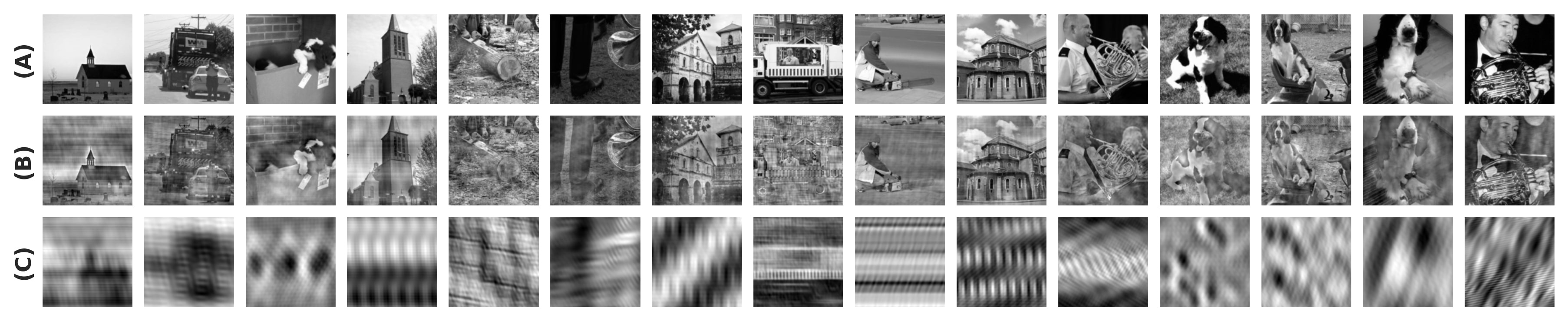}
\caption{(\emph{Top}) Randomly sampled single image masks divided by class. The colormap is the same as that in Figure \ref{fig:dataaug_vgg}.  (\emph{Bottom}) (A) randomly sampled images; (B) filtered by their complementary masks $M'_{N,x}$; (C) the same images filtered by their associated Fourier masks $M_{N, x}$.}
\label{fig:single_masks}
\end{figure}

To further investigate the nature of adversarial robustness and how it is related to a network's generalization properties in the frequency domain we generated Fourier masks $M_{N,x}$ for each single image $x$ in the validation set $X_{V}$. Figure \ref{fig:single_masks}(Top) shows such masks randomly sampled for images in all $5$ data classes calculated for the vanilla network $\Phi_{N}$. 

\begin{figure}[ht]\centering
\includegraphics[width=0.9\textwidth]{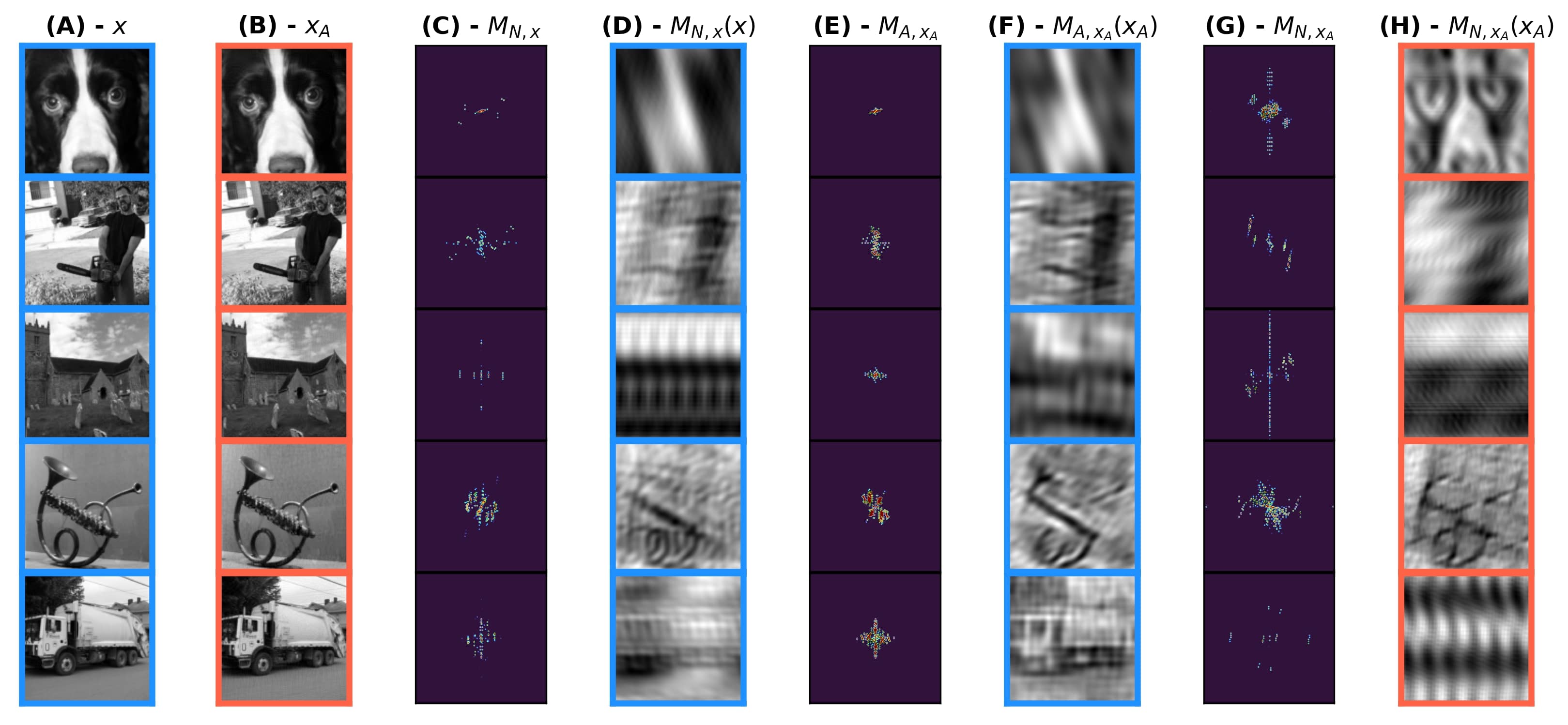}
\includegraphics[width=0.5\linewidth]{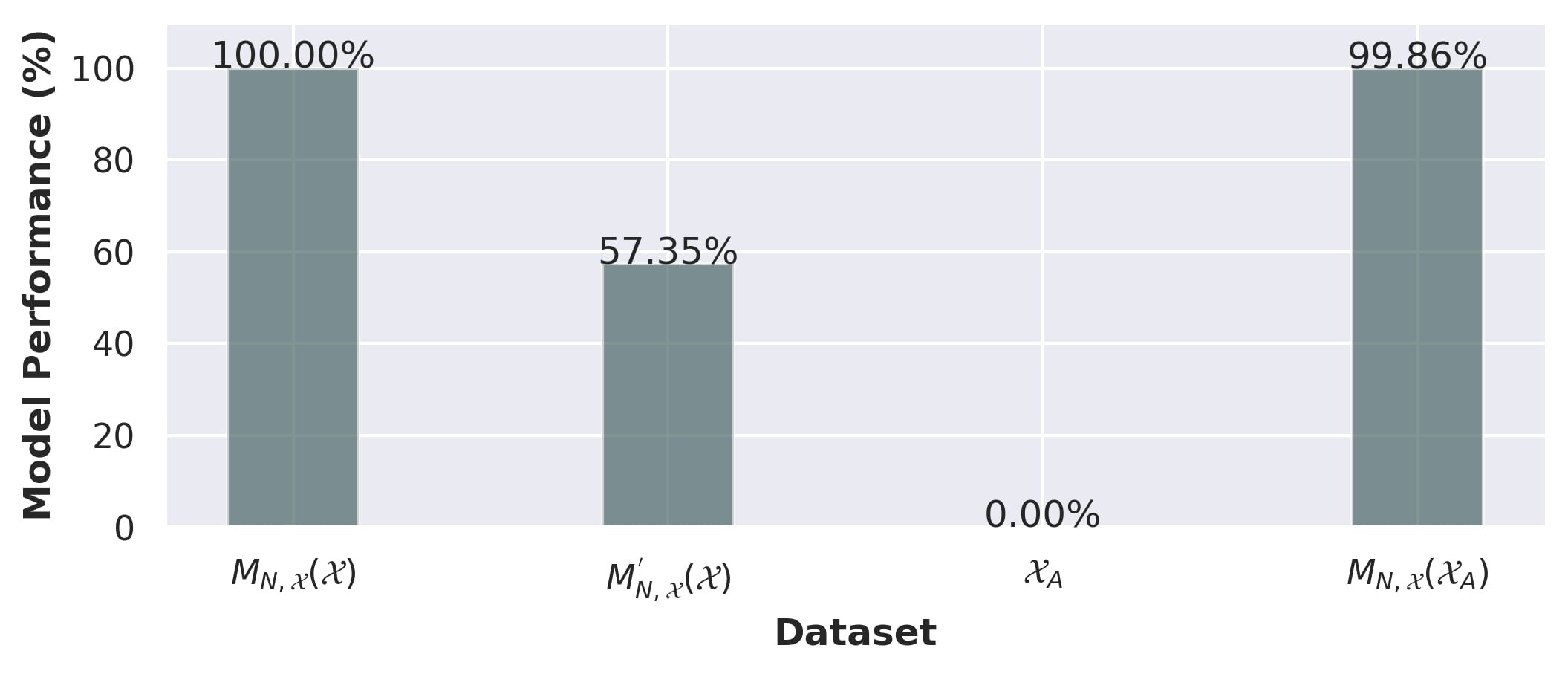}
\caption{(\emph{Top}): (A) Correctly classified image $x\in\mathcal{X}_{V}$ (Blue); (B) Adversarial Image $x_{A}$ (Orange); (C) $M_{N,x}$ - learned mask for vanilla network; (D) $M_{N,x}$ - filtered image $x$; (E) $M_{A,x_{A}}$ - learned mask for adversarially trained network; (F) $M_{A,x_{A}}$ - filtered adversarial image $x_{A}$; (G) $M_{N,x_{A}}$ - learned mask for vanilla network using the adversarial image $x_{A}$; (H) $M_{N,x_{A}}$ - filtered adversarial image $x_{A}$. The colormap is the same as that in Figure \ref{fig:dataaug_vgg}.
(\emph{Bottom}): Model performance on dataset $\mathcal{X}$ filtered by single-image vanilla masks $M_{N,\mathcal{X}}$; filtered by their complementary masks $M'_{N,\mathcal{X}}$; performance on the adversarial examples $\mathcal{X}_{A}$; and on the adversarial examples $\mathcal{X}_{A}$ filtered by single-image vanilla masks $M_{N,\mathcal{X}}$. We generate masks only for test images that the pretrained model classifies correctly.}
\label{fig:adv_single}
\end{figure}

It is worth noting that the masks are very sparse, i.e., very few frequencies are essential for preserving the prediction of the pretrained network. Additionally, for every mask $M_{N, x}$, we also consider its complementary mask $M'_{N, x}$ defined as
\[
M'_{N, x}(i, j) = \begin{cases}
1, & M_{N, x}(i, j) < 10^{-8} \\
0, & \text{otherwise.}
\end{cases}
\]
Filtering an image with its complementary mask $M'_{N, x}$ does not compromise our ability to recognize the filtered image (Figure \ref{fig:single_masks}(B, Bottom)). On the contrary, filtering with the mask $M_{N,x}$ renders the image unrecognizable (Figure \ref{fig:single_masks}(C, Bottom)). Filtered images resemble texture-like patterns. Interestingly, a recent work by [\cite{geirhos2018imagenet}] shows how  ImageNet-trained CNNs are strongly biased towards recognizing textures rather than shapes. Surprisingly, the opposite phenomenon is observed in the way the network perceives the changes induced by the respective masks. Performance drops drastically ($\sim 45\%$ decrease) for images filtered by complementary masks $M'_{N,x}$ Figure \ref{fig:adv_single}(Bottom). Additionally, filtering adversarial examples using masks generated from original images reverses the effect of the attack, restoring the original performance of the network. 

\begin{figure}[h!]\centering
\includegraphics[width=0.9\linewidth]{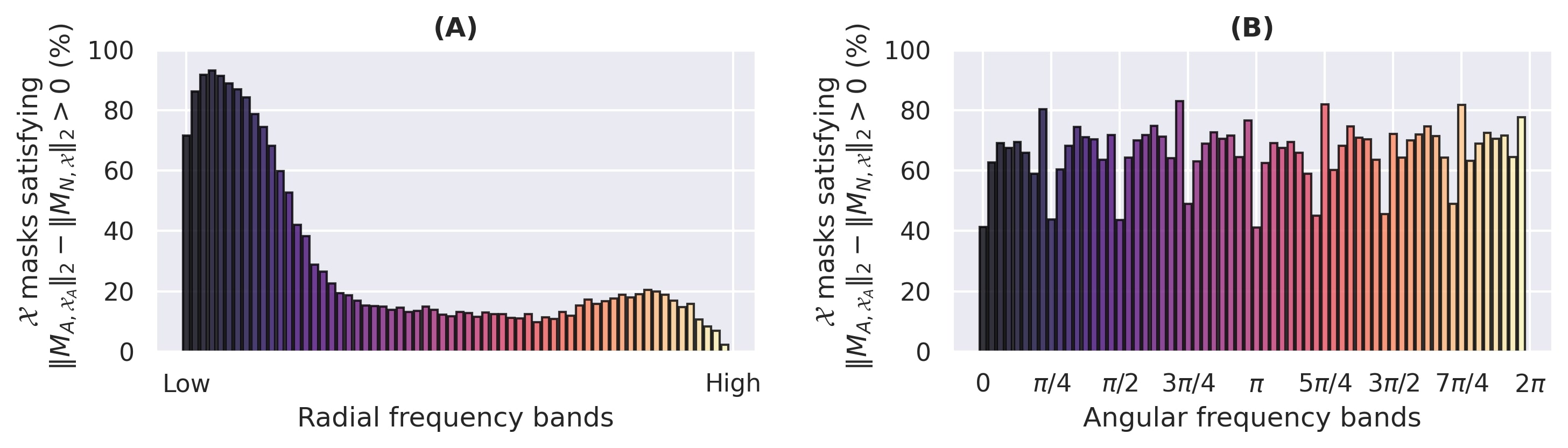}
\caption{Comparison of $\mathcal{X}$-trained masks ($M_{N,\mathcal{X}}$) obtained for the vanilla network and $\mathcal{X}_{A}$-trained masks ($M_{A,\mathcal{X}_{A}}$) obtained for an adversarially trained network: the bars illustrate the percentage of masks for which the per-band energy in $M_{A,\mathcal{X}_{A}}$ exceeds that of $M_{N,\mathcal{X}}$.}
\label{fig:adv_single_comparison}
\end{figure}

We also computed single image masks for clean images $x$ and adversarial images $x_{A}$ using either the naturally pretrained network $\Phi_{N}$ or the adversarially pretrained network $\Phi_{A}$. We compared the masks $M_{N,\mathcal{X}}$ with $M_{A,\mathcal{X}_{A}}$ to highlight the most important frequencies used by an adversarially trained network to make its predictions robust. We then compared the masks $M_{N,\mathcal{X}}$ with $M_{N,\mathcal{X}_{A}}$ to assess which frequencies are responsible for a network to make a wrong prediction on the adversarial image. Figure \ref{fig:adv_single} shows examples of such masks calculated for five randomly chosen validation images from each class.
We note that the masks $M_{A,\mathcal{X}_{A}}$ have more energy concentrated in lower frequencies compared to those in $M_{N,\mathcal{X}}$, further confirming a low frequency bias in adversarially trained networks. These observations are quantitatively substantiated by the plots in Fig. \ref{fig:adv_single_comparison} where we computed the percentage of perturbed images for which the per-band energy (radial or angular) of their corresponding masks $M_{A,\mathcal{X}_{A}}$ exceeds that of the masks $M_{N,\mathcal{X}}$ generated from the non-perturbed examples. Fig. \ref{fig:adv_single_comparison} confirms that lower frequencies are preferred for a robust representation.

\begin{figure}[h!]
\centering
\includegraphics[width=0.85\linewidth]{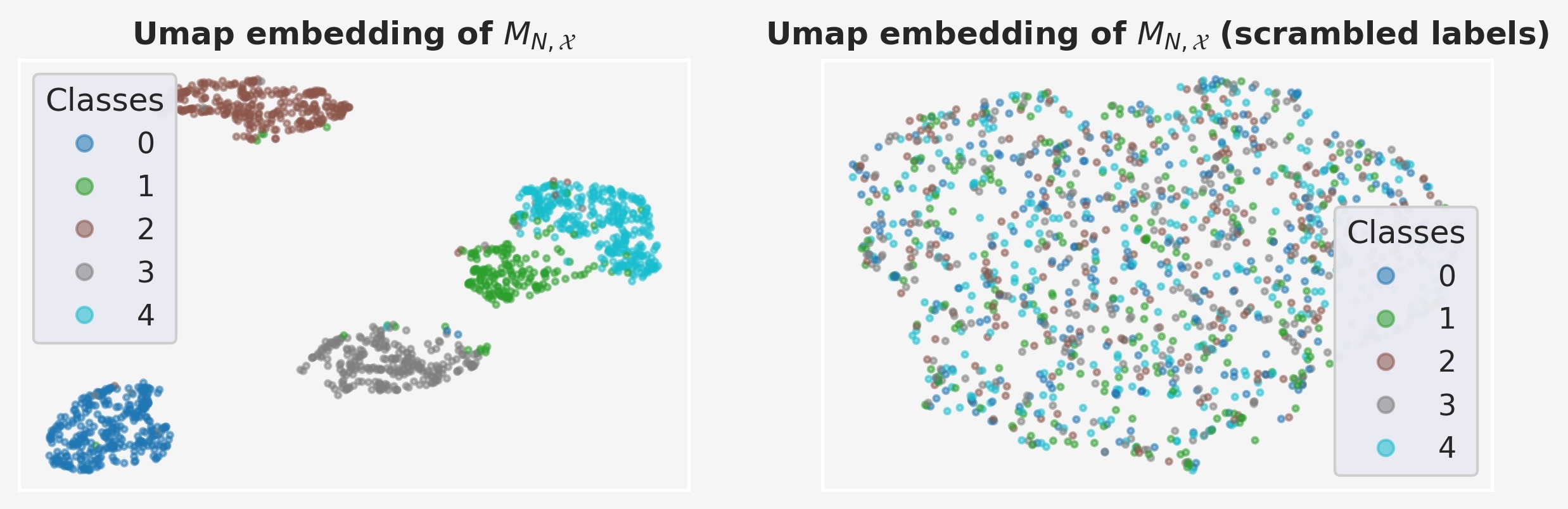}
\caption{Clustering of the output of a linear classifier learned to separate single image Fourier masks. $M_{N, \mathcal{X}}$ indicates the set of single image masks computed for the correctly classified test images in $\mathcal{X}$.}
\label{fig:umap}
\end{figure}

We then performed a manifold analysis of the learned masks. We found that the masks are linearly separable and that the linear network responses cluster (Figure \ref{fig:umap}).
At the same time, we see that this is not the case when the labels associated with the masks are shuffled. Therefore, the frequencies the network deems essential for prediction are class-specific since the results suggest that linear separability of the masks is due to their geometry and not the representation power of the linear classifier.

\section{Discussion and Conclusions}\label{sec:discussion}
In this work we proposed a simple yet powerful approach to visualize the essential frequencies a trained network is using to solve a task. Our strategy consists of learning a frequency modulatory mask characterized by two critical properties: 
\begin{itemize}
\item it defines a symmetry in the Cross Entropy loss, i.e., it does not alter the pretrained model's predictions.
\item it has minimal $\ell_{p}$-norm, which for $p=1$ guarantees the preservation of performance while promoting sparsity in the mask.
\end{itemize}
Using our method we tested the common hypothesis that adversarially trained networks prefer low frequency features to achieve robustness. We also tested if this hypothesis holds true for common data augmentations such as translations, scales, and rotations. 

In the case of adversarial augmentation, our results confirm the low frequency bias hypothesis. However, they also highlight that the frequency redistribution due to the augmentation is highly anisotropic. 
In the case of common data augmentations instead, our results show how the frequency reorganization depends on the type of augmentation.   

In the case of adversarial training we also run a single image analysis to detect the frequencies useful for adversarial robustness and those responsible for adversarial weakness. Here too, masks learned on adversarially trained networks concentrate more towards lower frequencies compared to those learned on vanilla networks. Furthermore, the analysis showed that only a sparse, class-specific set of frequencies is needed to classify an image. Surprisingly, mask-filtered images in this case are not recognizable and resemble texture-like patterns, supporting the idea that ANNs use fundamentally different classification strategies from humans to achieve robust generalization [\cite{Geirhos2019}].

To our knowledge the use of a learned mask to characterize a network's crucial property such as robust generalization has not been proposed before. The interpretation of the masks provides us with a detailed geometrical description of directional and radial biases in the frequency domain as well as with quantifiable differences between various training schemes.

Our analysis can be extended to other architectural or optimization specifics, e.g., explicit regularizations, different optimizers/initializations, etc. The same mask approach can be employed to modulate the phase and modulus in the Fourier transform of the data. Our method  effectively opens up many directions in the investigation of a network's implicit frequency bias. Future research directions will also include a natural generalization of our approach where the image features are learned, rather then fixed to be of the Fourier type.

\section*{Conflict of Interest Statement}
The authors declare that the research was conducted in the absence of any commercial or financial relationships that could be construed as a potential conflict of interest.

\section*{Author Contributions}
N.K. and F.A.  conceived the conceptualized framework. N.K. and E.B. trained and analyzed models. N.K. and F.A. wrote the first draft.  A.P., J.O.C., A.S.T. and X.P.  provided  feedback along the way. All authors revised, edited and provided comments on the final manuscript. A.S.T., A.P. and X.P. provided funding.

\section*{Availability of data and materials}
Code is available at https://github.com/nkarantzas/FourierMasks.git

\section*{Acknowledgments}
N.K., E.B., J.O.C., A.P., A.S.T., X.P. and  F.A.  acknowledges funding support from IARPA MICRONS, DARPA L2M, and NSF Neuronex.
We also thank Shell Xu Hu, Kandan Ramakrishnan, and Zhe Li for helpful discussions.

\bibliographystyle{unsrt}
\bibliography{Sym.bib}
\end{document}